\documentclass[runningheads]{llncs}
\usepackage[T1]{fontenc}
\usepackage{hyperref}
\usepackage{graphicx}
\usepackage{multirow}
\usepackage{adjustbox}
\usepackage{tabularx}
\usepackage{booktabs}
\usepackage{amsfonts}
\usepackage{pifont}
\usepackage{cite}
\usepackage{caption}
\begin{document}
\title{Anatomically Constrained Transformers for Cardiac Amyloidosis Classification}
% \begin{comment}  %% Removed for anonymized MICCAI 2025 submission
\author{Alexander Thorley\inst{1,7}
\and Agis Chartsias\inst{7}\and
Jordan Strom\inst{3} \and
Roberto Lang\inst{4} \and Jeremy Slivnick\inst{4} \and Jamie O'Driscoll\inst{5,6} \and Rajan Sharma\inst{5} \and Dipak Kotecha\inst{2} \and
Jinming Duan\inst{1} \and Alberto Gomez\inst{7}}
\authorrunning{A. Thorley et al.}
\institute{School of Computer Science, University of Birmingham, UK \email{ajt973@student.bham.ac.uk, j.duan@bham.ac.uk} \and
Institute of Cardiovascular Sciences, University of Birmingham, UK \and 
Beth Israel Deaconess Medical Center, Boston, Massachusetts, USA \and University of Chicago, USA \and St George’s University Hospitals NHS Foundation Trust, UK \and
Diabetes Research Centre, University of Leicester, UK
\and
Ultromics Ltd, Oxford, UK}
% \end{comment}

% \author{Anonymized Authors}
% \authorrunning{Anonymized Author et al.}
% \institute{Anonymized Affiliations \\
%     \email{email@anonymized.com}}
\maketitle 
\begin{abstract} Cardiac amyloidosis (CA) is a rare cardiomyopathy, with typical abnormalities in clinical measurements from echocardiograms such as reduced global longitudinal strain of the myocardium. An alternative approach for detecting CA is via neural networks, using video classification models such as convolutional neural networks. These models process entire video clips, but provide no assurance that classification is based on clinically relevant features known to be associated with CA. An alternative paradigm for disease classification is to apply models to quantitative features such as strain, ensuring that the classification relates to clinically relevant features. Drawing inspiration from this approach, we explicitly constrain a transformer model to the anatomical region where many known CA abnormalities occur- the myocardium, which we embed as a set of deforming points and corresponding sampled image patches into input tokens. We show that our anatomical constraint can also be applied to the popular self-supervised learning masked autoencoder pre-training, where we propose to mask and reconstruct only anatomical patches. We show that by constraining both the transformer and pre-training task to the myocardium where CA imaging features are localized, we achieve increased performance on a CA classification task compared to full video transformers. Our model provides an explicit guarantee that the classification is focused on only anatomical regions of the echo, and enables us to visualize transformer attention scores over the deforming myocardium.
\keywords{Transformers \and Echocardiography \and Pre-training}
\end{abstract}
\section{Introduction}
Echocardiography (echo) is commonly used for structural and functional assessment of the heart, including measurements of ejection fraction and longitudinal strain to support in the diagnosis of cardiac disease. In the case of cardiac amyloidosis (CA), a rare cardiomyopathy where a build up of amyloid fibrils occurs in the myocardium \cite{dorbala2021asnc}, abnormalities such as apical sparing are typical imaging features where longitudinal strain is reduced in the basal segments of the myocardium \cite{Sharmila}. However, such quantitative measures are prone to significant variability between different commercial software packages used to calculate them \cite{mirea2018variability}. Many works have instead presented deep learning techniques as an alternative for disease classification, mainly in the form of convolutional neural networks (CNNs) \cite{goto2021artificial} and more recently transformers \cite{mokhtari2023gemtrans, amadou2024echoapex}. These models are generally considered to be black boxes that process entire images or video clips and do not guarantee that the classification is based on clinically relevant image regions. In the pursuit of more explainable classification models, a number of prior works in echo \cite{sanchez2018machine,chiou2021ai, hathaway2022ultrasonic,upton2022automated,hughes2021deep,painchaud2024fusing} have constrained models to image-derived measurements and features, including strain and segmentation masks, to guarantee that models are focused on clinically relevant features. In this paper, we draw inspiration from these prior works and approach the CA classification task with a model constrained to the myocardium, where many known CA imaging features are located. We intend to incorporate anatomical information about the myocardium, which we parameterize as a set of points deforming through an echo, since structural and functional metrics such as increased wall thickness and apical sparing are common in CA patients \cite{Sharmila}. We also aim to incorporate the underlying image information from the region as a ``sparkling, hyper-refractile texture of the myocardium''\cite{Sharmila} is another known CA feature. Our model is designed to incorporate both structural and textual information localized to the myocardium, enabling us to process regions of differing shapes and sizes depending on the patient.
\\
\indent
The transformer \cite{vaswani2017attention}
naturally lends itself to this problem as it does not assume that input data follows a particular structure, providing the required flexibility to accommodate varying myocardium sizes. % us to model myocardiums of differing shapes and sizes. %The transformer has seen recent success in the domain of echocardiography in the form of vision transformers (ViTs).
 Related works in echo have typically used variants of the vision transformers (ViTs) \cite{vaswani2017attention}, with the space-time factorised ViViT \cite{arnab2021vivit} proving most popular. The model encodes each frame independently with a ViT, and aggregates the resulting class tokens encodings with a temporal transformer \cite{mokhtari2023gemtrans,amadou2024echoapex,fadnavis2024echofm}. Alternatively, entire video clips can be processed with a single video transformer \cite{szijarto2024masked}. To achieve sufficient performance with transformers, part or all of such models are first pre-trained via a self-supervised learning (SSL) with methods such as MAE \cite{he2022masked} or DINO \cite{caron2021emerging} to learn rich representations from unlabeled data. Pre-trained models can then be tuned on a smaller set of labeled images for a wide range of tasks including segmentation and classification. In this paper, we apply an anatomical constraint to both the transformer and pre-training task itself. We note that work from \cite{szijarto2024masked} introduced a transformer and pre-training task constrained to only the ultrasound triangle region of interest (ROI), and \cite{mokhtari2023gemtrans} applied an additional loss to their ViT to encourage attention weights to focus on anatomically relevant regions of the echo. However, neither works \emph{explicitly} constrain the transformer to an anatomical region of interest, which we tackle with the following contributions: 
\begin{itemize}
    \item We propose an anatomically constrained transformer for echo video processing (ViACT), an adaptation of the space-time factorized ViViT \cite{arnab2021vivit} where our tokenizer embeds myocardium points and corresponding sampled image patches. This explicitly constrains the model to regions covered by the deforming points through the video, guaranteeing that the CA classification is derived from image regions with known clinical indicators of CA.
    \item We introduce an anatomical variation of the MAE framework \cite{he2022masked} to pre-train the ViACT frame encoder by masking and reconstructing myocardium patches with embedded point locations as positional embeddings in both the encoder and decoder. When tuned for CA classification, we show that the combination of the anatomically constrained model and pre-training demonstrates improved performance over conventional image based transformers.
    % \item We show that for our application of CA classifcation, this explicitly ensures the model is focused on the myocardium where apical sparing and increased wall thickness are known clinical indicators of the disease
    % \cite{dorbala2021asnc} and results in improved performance against models processing the full ultrasound images.
\end{itemize}
\section{Methodology}
\label{sec:methodology}
\begin{figure}[t!]
\includegraphics[width=\textwidth]{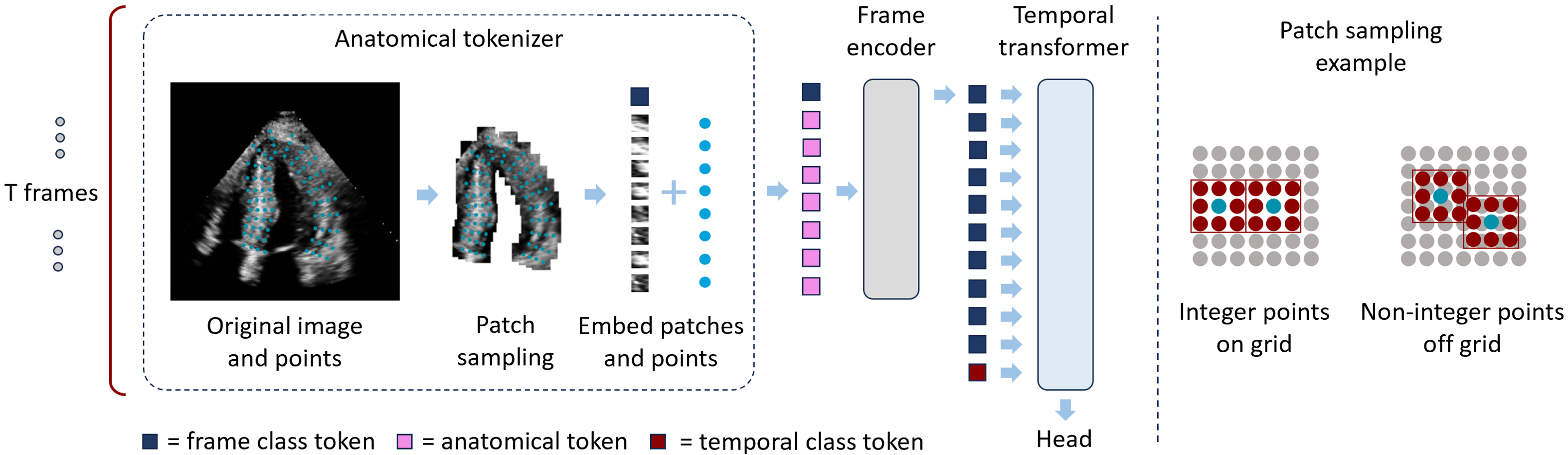} 
\caption{Left: the ViACT model comprised of an anatomical tokenizer, frame encoder and temporal transformer. Right: example $3 \times 3$ patches (red) centered at integer and non-integer points (teal) over a grid of pixels (gray).} \label{fig:main_pipeline}
\end{figure}

\begin{figure}[t!]
\includegraphics[width=\textwidth]{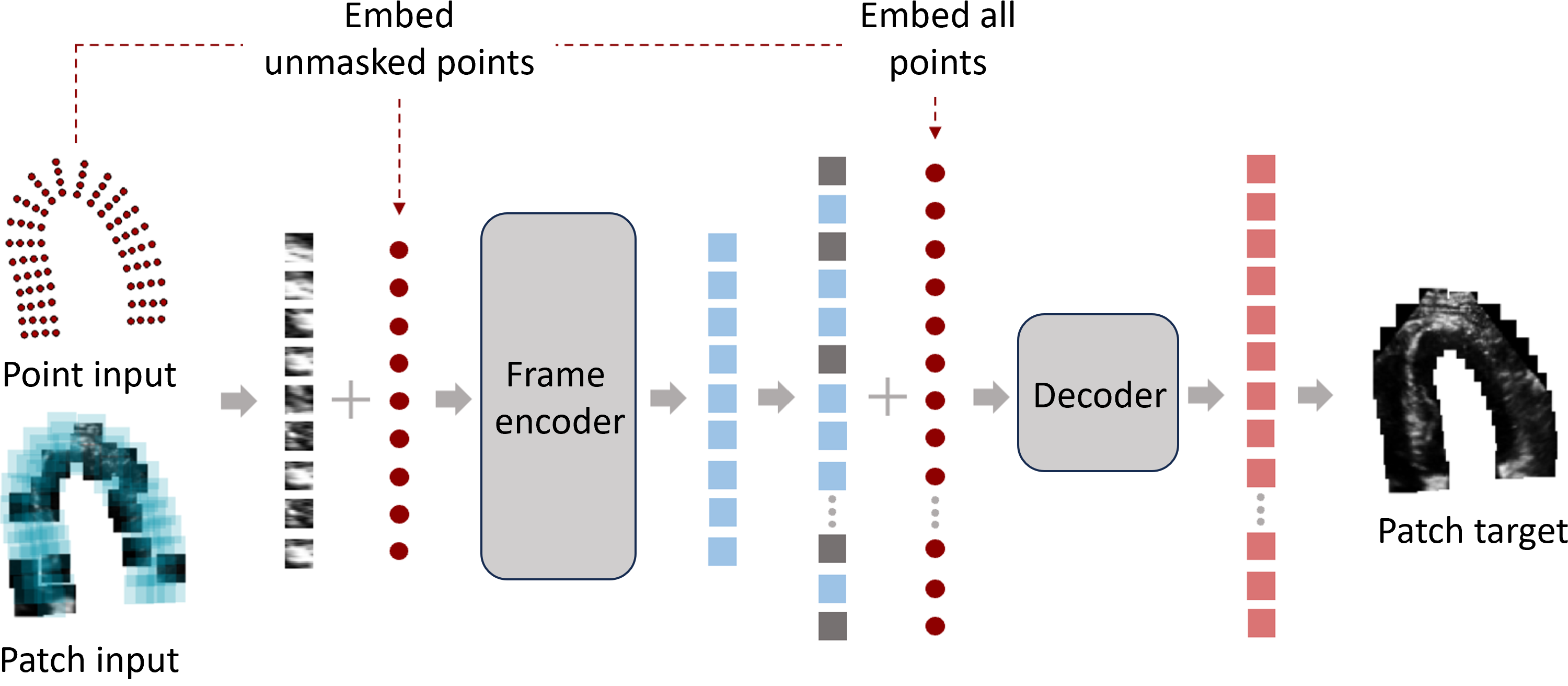} 
\caption{The anatomical MAE pipeline for the ViACT frame encoder, with point embeddings added to input tokens of both the encoder and decoder. Masked patches are depicted with transparent blue squares. Graphic inspired by \cite{he2022masked}.} \label{fig:masking_strategies}
\end{figure}

\begin{figure}[t!]
\includegraphics[width=\textwidth]{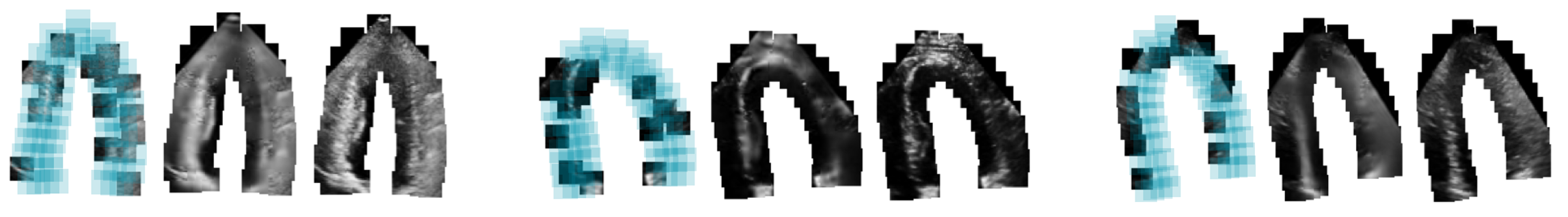} 
\caption{Example masked myocardium patches, reconstructions, and original patches from anatomical MAE pre-training.} \label{fig:reconstructions}
\end{figure}

In this section we introduce our anatomical tokenizer, provide a brief summary of the ViViT style \cite{arnab2021vivit} transformer backbone and detail our anatomical MAE pre-training strategy. Let us first consider ultrasound videos $V = (I_0, I_1, ... , I_T)$ composed of frames $I_t \in {\mathbb{R}^{H W}}$, where $t \in [0, T], T \in \mathbb{N}$. In this paper, we assume we have obtained corresponding sequences of points $(P_t^i)_{i=0, t=0}^{N, T}$ where each point $P_t^i = (x_t^i, y_t^i) \in {\mathbb{R}^{2}}$ denotes the $i^{th}$ point of a point set covering the myocardium in the $t^{th}$ frame, where $i \in [0, N], N \in \mathbb{N}$. This paper focuses on adapting a transformer to process both the video sequence and corresponding points rather than the acquisition of the points themselves. Details of how we obtained points for our experiments are found in section \ref{sec:experimental_results}. The complete model architecture with an example set of myocardium points for a single frame can be found in figure \ref{fig:main_pipeline}, left, and we detail each of the model components below.
% \begin{figure} 
% \includegraphics[width=\textwidth]{Images/pipeline.PNG} 
% \caption{The ViACT adaption of a ViVit style space-time factorized transformer, comprised of an ACT frame encoder and temporal transformer. Navy blue tokens represent frame class tokens and the red token corresponds to the temporal class token.} \label{fig:main_pipeline}
% \end{figure}
\\
\indent
\textbf{Anatomical tokenizer:} The regular ViViT uses a standard ViT frame encoder, where each image frame $I_t$ is split into patches on a regular integer grid, which is often implemented with a 2D convolution in practice. Patches are then linearly embedded and summed with a \emph{sin cos} or learnable positional embedding to create input tokens \cite{dosovitskiy2020image}. In contrast, our anatomical tokenizer is designed to embed each myocardium point $P_t^i$ and the patch of pixel information underneath it. Each of the myocardium points $P_t^i$ are non-integer coordinate locations rather than integer pixel indices and as the myocardium points deform over time there may exist an overlap between patches as points move closer together. An example showing the difference between extracting patches at integer locations (as per a regular ViT tokenizer) and non integer points can be found in figure\ref{fig:main_pipeline}, right. One can see that in the non-integer case, pixel intensity values must be interpolated in order to extract a patch centered on the points as they do not align with integer pixel indices. To do this, we constructed $j \times j$ sampling grids centered at each point $P_t^i$, where $j \in \mathbb{N}$ is the intended patch size of the tokens. Frame $I_t$ was then sampled via bilinear interpolation with the sampling grids to extract patch $C_t^i \in {\mathbb{R}^{j j}}$. Each sampled patch was subsequently flattened and linearly embedded to a vector of length $k$ corresponding to the transformer embedding dimension, where $k \in \mathbb{N}$. For our positional embeddings, we instead used linear projections of points $(P_t^i)_{i=0}^N$ to vectors of length $k$, which were summed with the embedded patches to form input tokens $(\alpha_t^i)_{i=0}^{N}$ where $\alpha_t^i \in {\mathbb{R}^{k}}$. This enabled us to embed myocardium points and underlying patches into tokens whilst assuming no order or structure to the myocardium point set. As such, the model can adapt flexibly to cardiac anatomy of any shape and size. 
% As we embed spatial location rather than position in a sequence, we can in theory handle myocardium meshes with varying numbers of points depending on the size and shape of the patient's heart. We do however keep the number of points fixed in our experiments for simplicity.
\\
\indent
\textbf{Transformer:} The tokenized myocardium points and patches are then fed into a space-time factorised ViViT-style transformer \cite{arnab2021vivit}. In brief, the frame encoder is a standard transformer encoder \cite{vaswani2017attention} which processes the myocardium tokens with an appended frame class token $\theta_t \in \mathbb{R}^k$. This is repeated for each of the $T$ frames resulting in $T$ class token encodings $(\hat{\theta}_t)_{t=0}^T$, where $\hat{\theta}_t \in \mathbb{R}^k$. A learnable positional embedding was then added to each $(\hat{\theta}_t)_{t=0}^T$, which were then fed to a temporal transformer with a temporal class token $\omega \in \mathbb{R}^k$. A linear classification head was attached to the corresponding encoded class token $\hat{\omega} \in \mathbb{R}^k$ and trained with a standard binary cross entropy loss to classify CA.
% \begin{figure} 
% \includegraphics[width=\textwidth]{Images/mae_pipeline.PNG} 
% \caption{The adapted anatomical MAE pipeline with point embeddings added to input tokens of both the encoder and decoder. Graphic inspired by \cite{he2022masked}.} \label{fig:masking_strategies}
% \end{figure}
\\
\indent
 \textbf{Anatomical MAE:} Pre-training has been shown as an essential step to achieve sufficient performance with transformers \cite{amadou2024echoapex,fadnavis2024echofm,kim2024echofm}. However, image pre-training strategies for ViT frame encoders such as MAE \cite{he2022masked} and DINO \cite{darcet2023vision} operate on a full grid of image patches, which is not compatible with the tokenizer and frame encoder of our ViACT model. Where the original MAE pipeline \cite{he2022masked} splits an image into patches on an integer grid and masks a proportion of these patches, we instead mask a subset of our myocardium patches and points. The remainder are embedded with our anatomical tokenizer. The frame encoder then processes this subset of anatomical tokens, whose encoded tokens are then padded out with mask tokens to align with the number and positions of the original patches and points. In contrast to the original MAE recipe \cite{he2022masked}, we added a linear projection of the original point location to each masked and unmasked token as positional embeddings. The coordinate location of each patch is therefore encoded into the input tokens of both the encoder and decoder. The full set of tokens is then processed by a small transformer decoder with a reconstruction head tasked with reconstructing the myocardium image patches under a mean squared error (MSE) loss between only the predicted and masked myocardium patches. For tuning, the decoder is discarded and the encoder is used as the frame encoder and tuned using all input tokens for a CA classification. Our adjustments to the MAE pre-training pipeline are shown in figure \ref{fig:masking_strategies}, with example reconstructions shown in figure \ref{fig:reconstructions}.
% \begin{figure} 
% \includegraphics[width=\textwidth]{Images/reconstructions.PNG} 
% \caption{Example masked myocardium patches, reconstructions, and original patches. Masked patches are depicted with transparent blue squares.} \label{fig:reconstructions}
% \end{figure}
 \\
 \indent
\section{Experiments and Results}
\label{sec:experimental_results}
\begin{figure}[b!]
\includegraphics[width=\textwidth]{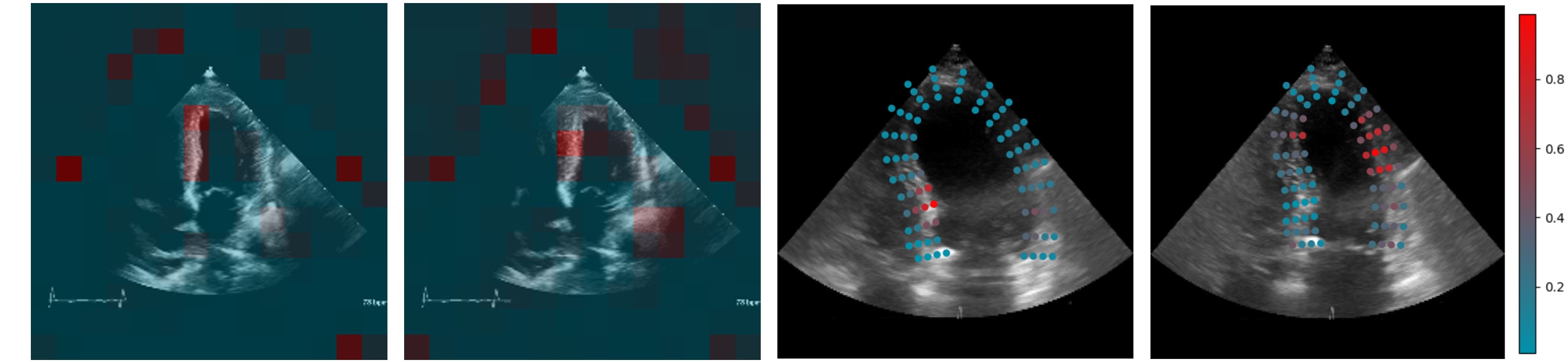} 
\caption{Left: attention maps from a ViViT. Right: points from a ViACT, colored with attention map. Attention maps are from a single head belonging to the class token of the final frame encoder transformer block and normalized.} \label{fig:attention_sequence}
\end{figure}

% Train: 306 case, 1066 control
% Val: 64 case, 230 control
% Test: 69 case, 224 control 
% Pre-training: 89,485
Experiments were conducted on a private, multi-site dataset comprised of 1959 4-chamber echocardiograms in DICOM format, of which 1520 were in the control group and 439 were CA cases comprised of both \emph{amyloid transthyretin} (ATTR) and \emph{amyloid light-chain} (AL). CA cases were determined according to the guidelines in \cite{Sharmila}. The control group was a mix of patients with obstructive and non-obstructive hypertrophic cardiomyopathy, hypertension with left ventricular hypertrophy, heart failure, suspected CA patients excluded via a Tc-PyP scan and healthy controls. Both CA patients and 
controls were identified via a retrospective review of participating sites databases, organized by the University of Chicago. All models in our results were bench-marked on a binary CA classification task.
\\
\indent
\textbf{Pre-processing:} Video clips were extracted from DICOM files and any pixel information beyond the ultrasound triangle was masked. Frames were gray-scaled, normalized to a mean and standard deviation of 0.5, and resized to $224\times224$ pixels. To extract points $(P_t^i)_{i=0, t=0}^{N, T}$ for our experiments we processed the DICOMs with a commercially available automated contouring software package EchoGo Core used in \cite{asch2022human}, producing 21 endocardium contour points for each frame. Such points could however be readily replaced with any other point extraction and tracking methods such as \cite{chernyshov2024automated,azad2024echotracker}. To ensure the model tokenized the entire myocardium, we spread three additional rows of points perpendicular to the contour at intervals of six pixels, resulting in a fixed number of 84 points per frame. Both image frames and contour points were resampled to a constant frame-time of 33.33ms. We selected sequences of 18 frames and points for all patients in our dataset starting from an end diastole frame. Whilst we fixed the number of points and clip lengths for simplicity, the model is not limited to a specific number of points or frames. We opted for a 70/15/15 split for our train/val/test sets, resulting in 1372, 294 and 293 samples, respectively, with each sample from a unique patient. We ensured that the prevalence of CA cases across the three sets was approximately equal. For pre-training, we used the full length clip and point sequences for patients in the training set. This resulted in a total of 89,485 individual frames and corresponding point sets.
\begin{figure}[t!]
\includegraphics[width=\textwidth]{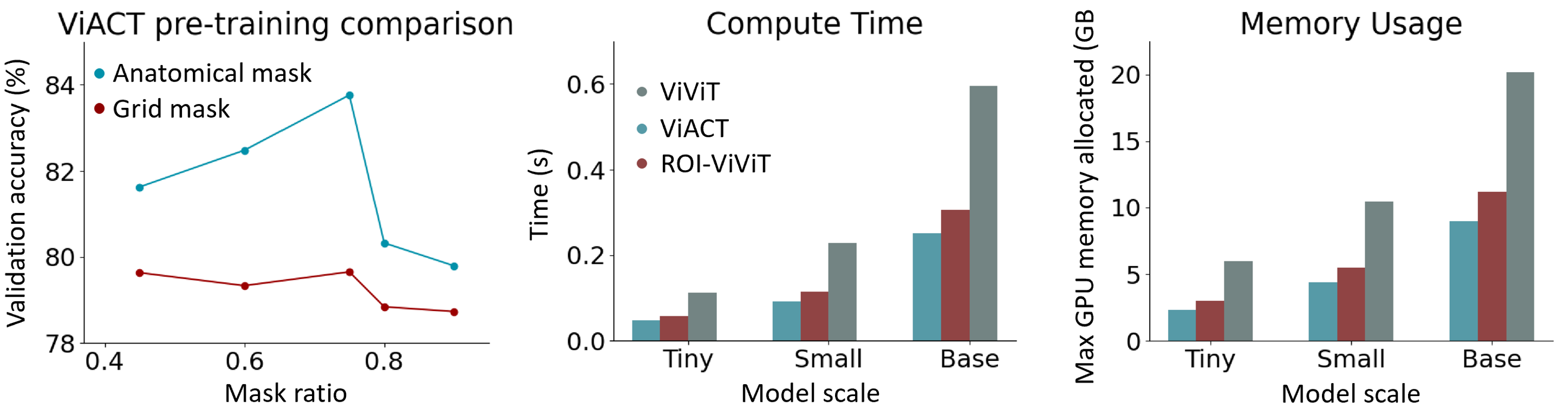} 
\caption{(1) Validation accuracy for "tiny" scale ViACT with different mask ratios and pre-training. (2) Memory usage and (3) compute time for a pre-training forward and backward pass for different model sizes with a batch of 400.} \label{fig:ablation_graphs}
\end{figure}
\\
\indent
\textbf{Model hyperparameters:} All our experiments used a ``tiny scale'' transformer for frame encoders with a patch size of 16. Across all models, transformer blocks used 3 heads with an embedding dimension of 192. Encoders were comprised of 12 transformer blocks, decoders for pre-training contained 4 transformer blocks  and temporal transformers used a single transformer block to process the fixed 18 frames with learnable positional embeddings. We used standard transformer blocks described in \cite{dosovitskiy2020image}, composed of multi head self-attention followed by an MLP, with layernorm before and residual connection after each. The MLP hidden feature dimension was 768 and we used GELU activation functions. Classification heads were comprised of a single linear layer from the embedding dimension to the binary classes.
\\
\indent
\textbf{Training configuration:} we followed the original MAE \cite{he2022masked} pipeline, scaled down to our dataset size. All models were trained with an AdamW optimiser \cite{loshchilov2017decoupled} using momentum of (0.9, 0.999) and weight decay of 0.05. For pre-training, we used a learning rate scheduler comprised of a linear warm up of 200 epochs followed by cosine decay \cite{loshchilov2016sgdr} for 1800 epochs. No scheduler was used for tuning. Base learning rates were scaled by (batch size / 256) as per \cite{goyal2017accurate} using batch sizes of 2700 and 35 with base learning rates of 1.5e-4 and 1e-3  for pre-training and tuning, respectively. All experiments were distributed over two NVIDIA GeForce RTX 3090 24.5G GPUs and implemented in PyTorch 3.1. Pre-training was run only once per model. We stopped tuning when the best validation loss failed to improve after 8 epochs, selecting the model with the best validation loss and repeating runs 50 times with differing seeds. We used accuracy and a weighted F1 score as our performance measures averaged over all repeats.
\\
\indent
\textbf{Ablation study:} to establish the importance of the anatomical MAE pre-training, we pre-trained two ViACT frame encoders- one using anatomical MAE and one on a grid of points and patches covering the entire image instead. We varied the pre-training mask ratio (the ratio of masked to total tokens) from 45-90\% for each model and tuned the pretrained ViACTs on myocardium tokens for CA classification. This enabled us to determine if pre-training on the myocardium patches offered a benefit over using all image patches. As shown in figure \ref{fig:ablation_graphs}, plot 1, we observed peak performance on the validation set at a mask ratio of 75\% for both variants, with the anatomical MAE demonstrating a 4\% point improvement in accuracy. We also pre-trained and tuned models using a variety of different positional embeddings to determine the importance of this design choice. We compared the \emph{point linear projection} embeddings described in section \ref{sec:methodology} with a \emph{point sin cos} variant, where a \emph{sin cos} embedding was applied to \emph{x} and \emph{y} coordinates and averaged. We also experimented with an \emph{apex relative} version, where each point was embedded relative to the apex point of the set as $\rho(P_t^i - P_t^{apex})$, where $\rho$ is either an averaged \emph{sin cos} embedding or \emph{linear} projection. This version is invariant to the position of the myocardium points in the image, however we found performance on the validation set to be greater with a simple linear projection of the point locations as seen in table \ref{table:ablation_study}.
\begin{table}[t!]
\centering
\fontsize{8}{10}\selectfont
\caption{Ablation study of positional embeddings and pre-training on validation set. Results averaged over fifty repeats, showing mean $\pm$ standard deviation.}
\begin{tabular}{|l|l|l|l|l|l}
\hline
Positional embedding & MAE pre-training & Accuracy & Weighted F1 \\
\hline
Apex relative sin cos & Anatomical & 79.54 $\pm$ 1.49 & 75.61 $\pm$ 4.38
\\
Apex relative linear & Anatomical & 80.26 $\pm$ 1.64 &  76.29 $\pm$ 4.11
\\
% ViACT & Anatomical MAE & 75\% & Learnable & 80.18 $\pm$ 2.50 & 78.59 $\pm$ 4.14
% \\
Point sin cos & Anatomical & 81.65 $\pm$ 1.06 & 79.74 $\pm$ 2.01
\\
Point linear projection & None & 80.24 $\pm$ 1.41 & 78.17 $\pm$ 2.59
\\
Point linear projection & Grid & 79.65 $\pm$ 1.53 & 74.99 $\pm$ 3.62
\\
Point linear projection & Anatomical & \textbf{83.76 $\pm$ 1.25} & \textbf{82.62 $\pm$ 1.65}
\\
\hline
\end{tabular}
\label{table:ablation_study}
\end{table}
 \\
\indent
\textbf{Results:} We evaluated our ViACT against the space-time factorized ViViT, which is popular in the literature for echo processing tasks \cite{amadou2024echoapex,mokhtari2023gemtrans,fadnavis2024echofm}. We pre-trained the frame encoder with MAE \cite{he2022masked}. Work from  \cite{szijarto2024masked} applied the VideoMAE2 framework to only patches contained within the ultrasound ROI- we compared with a space-time factorized ViViT constrained the ROI. ROIs for all training clips were calculated using the approach from \cite{szijarto2024masked}, and the average ROI from training samples was used to constrain the model. We both pre-trained the frame encoder with MAE and inferred on only ROI patches, enabling us to directly compare the performance of a ROI model against our anatomical version with regular MAE pre-training and an identical space-time factorized backbone. Finally, we compared with a spatio-temporal MAE\cite{feichtenhofer2022masked} (MAE-ST), pre-trained with patch size of $1\times16\times16$. Results for all models can be found in table \ref{table:experimental_results}, where it can be seen that our ViACT model outperforms the other methods in terms of both accuracy and weighted F1 score. Furthermore, as seen in figure \ref{fig:ablation_graphs} the ViACT also used less than half the GPU memory and compute time for pre-training compared with a regular spacetime-factorized ViViT due to the large reduction of input tokens. This is of great importance for scaling to larger model and dataset sizes where pre-training GPU requirements become prohibitive. We note that there is an additional cost to obtain the tracked contours which will vary depend on the method used, however this is done once offline prior to pre-training. Compared with a standard ViViT, tokens and attention maps from the ViACT model are guaranteed to be focused on the myocardium providing a safeguard that the model is only using anatomical pixel information to make a classification. An example case seen in figure \ref{fig:attention_sequence} shows higher attention score's clustering around localized regions of the myocardium.
\begin{table}[t!]
\centering
\fontsize{8}{10}\selectfont
\caption{Test set results for comparative methods.}
\begin{tabular}{|l|l|l|l|l|l}
\hline
Model & MAE pre-training & Accuracy & Weighted F1 \\
\hline
ViVit & Grid & 78.08 $\pm$ 2.41 & 75.98 $\pm$ 3.69 \\
ROI-ViVit & ROI & 78.03 $\pm$ 2.93 & 75.82 $\pm$ 3.73 
\\
MAE-ST & Spatio-temporal & 80.25 $\pm$ 1.65 & 76.87 $\pm$ 3.24
\\
ViACT & Anatomical & \textbf{81.58 $\pm$ 1.70} & \textbf{80.69 $\pm$ 1.74}
\\
\hline
\end{tabular}
\label{table:experimental_results}
\end{table}
\section{Conclusion}
In this paper, we have presented a ViACT model pre-trained with anatomical MAE to process patches and points constrained to the myocardium for CA classification. The model outperformed standard ViVit models on a small scale CA dataset, whilst dramatically reducing pre-training times and producing attention maps focused on the myocardium where CA is known to present abnormalities. This work sets the stage for future research scaling dataset and model size, and leveraging the capacity of the model to process patient specific point sets with varying numbers of points and cycle lengths.
% \begin{comment}  %% removed for anonymized MICCAI 2025 submission.
    
%     % The following acknowledgement and disclaimer sections should be removed for the double-blind review process.  
%     % If and when your paper is accepted, reinsert the acknowledgement and the disclaimer clause in your final camera-ready version.

% \begin{credits}
% \subsubsection{\ackname} A bold run-in heading in small font size at the end of the paper is
% used for general acknowledgments, for example: This study was funded

% \subsubsection{\discintname}
% It is now necessary to declare any competing interests or to specifically
% state that the authors have no competing interests. Please place the
% statement with a bold run-in heading in small font size beneath the
% (optional) acknowledgments\footnote{If EquinOCS, our proceedings submission
% system, is used, then the disclaimer can be provided directly in the system.},
% for example: The authors have no competing interests to declare that are
% relevant to the content of this article. Or: Author A has received research
% grants from Company W. Author B has received a speaker honorarium from
% Company X and owns stock in Company Y. Author C is a member of committee Z.
% \end{credits}

% \end{comment}

\bibliographystyle{splncs04}
\bibliography{ref}

\end{document}